\documentclass[runningheads]{llncs}
\pdfoutput=1
\usepackage{booktabs}   % added by Sean
\usepackage{wrapfig}    % added by Sean
\usepackage{subcaption} % added by Sean
\usepackage[hidelinks]{hyperref} 
\usepackage{amsmath}
\DeclareMathOperator*{\argmax}{arg\,max}
\usepackage{tabularx,array}
\newcolumntype{Y}{>{\centering\arraybackslash}X}
\usepackage[T1]{fontenc}
\usepackage{graphicx}
\begin{document}
\title{Interpretable Tile-Based Classification of Paclitaxel Exposure}


\author{Sean Fletcher\inst{1} \and
Gabby Scott\inst{1}\and
Douglas Currie\inst{1}\and
Xin Zhang\inst{1}\and
Yuqi Song\inst{1}\and
Bruce MacLeod\inst{1}
}
\authorrunning{S. Fletcher et al.}

\institute{University of Southern Maine, Portland, United States 
\email{\{sean.fletcher,gabrielle.scott1,douglas.currie,xin.zhang,yuqi.song\}@maine.edu}}

\maketitle              % typeset the header of the contribution
\begin{abstract}
Medical image analysis is central to drug discovery and preclinical evaluation, where scalable, objective readouts can accelerate decision-making. We address classification of paclitaxel (Taxol) exposure from phase-contrast microscopy of C6 glioma cells—a task with subtle dose differences that challenges full-image models. We propose a simple tiling-and-aggregation pipeline that operates on local patches and combines tile outputs into an image label, achieving state-of-the-art accuracy on the benchmark dataset and improving over the published baseline by ~20 percentage points, with trends confirmed by cross-validation. To understand why tiling is effective, we further apply Grad-CAM/Score-CAM and attention analyses, which enhance model interpretability and point toward robustness-oriented directions for future medical image research. Code is released to facilitate reproduction and extension.

\end{abstract}

\section{Introduction}
Medical image diagnosis is fundamental to biomedical research and clinical decision making~\cite{Litjens2017Survey}. Deep learning–assisted analysis can reduce manual workload, improve reproducibility, and enable studies at scales impractical for human inspection~\cite{Esteva2017Derm}. Within this agenda, we focus on paclitaxel (paclitaxel, commonly known as Taxol), an anti-mitotic chemotherapeutic whose effects appear as subtle morphological changes in cell images. Classifying images by exposure level provides a concrete benchmark for drug-response phenotyping and can inform dose–efficacy and dose–toxicity assessment.

In this application area, a Taxol-response image dataset and a baseline coupling a CNN encoder with a k-nearest neighbors (kNN) classifier in the learned embedding space were introduced in~\cite{fletcher2025taxol}. While this configuration is standard in microscopy analysis, its accuracy on the Taxol task remains limited, reflecting the difficulty of separating adjacent exposure levels when full-image training dilutes fine-scale cues and image context varies across plates or sessions.

Motivated by this gap, we introduce tiling as a drop-in enhancement to the baseline~\cite{Sharma2024,Vacca2024}. Each image is partitioned into a regular grid of patches for representation learning; at inference, patch-level predictions are aggregated into a single image label using majority or probability-weighted voting. This design preserves the original backbone and evaluators (KNN and a conventional fully connected head) while emphasizing localized structure that would otherwise be averaged out at full-image scale. Across grid sizes, tiling yields consistent gains, with fine-grained splits and probability-weighted aggregation achieving the highest test accuracy.

To understand why tiling improves performance, we combine quantitative evaluation with interpretability based on class-activation heat maps~\cite{selvaraju2017grad,wang2020score}.The visualizations suggest that tiling encourages the model to attend to localized, fine-scale structures distributed across the field—such as peri-cellular details and culture-context microtextures—rather than relying solely on prominent cell contours. These contextual cues vary with paclitaxel dose and are consistent within tiles; aggregating tile-level predictions with probability-weighted voting acts as an ensemble that stabilizes image-level decisions and yields consistent gains. Moreover, this observation also points to robustness-focused directions for future microscopy-based cell classification. Because background and imaging conditions—such as plate artifacts, instrument differences, and illumination—vary more across experiments than the underlying cellular morphology, models that rely on these non-cellular cues are especially prone to distribution shift and exhibit reduced transferability.

This work makes the following contributions:
\begin{itemize}
\item We introduce a tiling-based enhancement to a standard CNN+KNN baseline to improve Taxol classification accuracy over full-image training.
\item We conduct an interpretability study using attention mechanisms and heat-map visualizations to explain that the performance gains arise from localized, context-aware image statistics associated with paclitaxel exposure.
\item We make our implementation open-source to enable reproduction and further research on drug-response phenotyping\footnote{\url{https://anonymous.4open.science/r/Microscopy_Image_Classification-BA26/README_overview.md}}.
\end{itemize}

\section{Related Work}
\subsection{ResNet-50 in Histopathology}

ResNet-50, a residual convolutional neural network architecture introduced by He et~al.~\cite{he2016deep}, has become a widely used backbone in histopathological image analysis due to its strong feature extraction capabilities, efficient optimization, and adaptability to transfer learning workflows. Its applications span multiple tissue types and diagnostic tasks, with reported success in both patch-level and whole-slide classification.

In breast histology classification, Vesal et~al.\ fine-tuned ImageNet-pretrained ResNet-50 to categorize image patches from the BACH 2018 challenge into four classes—normal, benign, \textit{in situ} carcinoma, and invasive carcinoma—achieving 97.5\% accuracy and outperforming an Inception-V3 baseline~\cite{Vesal2018}. Similarly, Talo~cite{Talo2019} applied transfer learning with ResNet-50 and DenseNet-161 to digital histopathology patches, obtaining up to 98.9\% accuracy for color images and demonstrating strong performance even on grayscale inputs. Mahmud et~al.~\cite{Mahmud2023} evaluated ResNet-50 alongside deeper variants (ResNet-101, VGG16/19) on the invasive ductal carcinoma (IDC) dataset, with ResNet-50 achieving the highest overall accuracy (90.2\%), AUC (90.0\%), and recall (94.7\%). Beyond breast tissue, Emegano et~al.~\cite{Emegano2025} used ResNet-50 for prostate cancer histopathology classification, underscoring the architecture’s adaptability across cancer types. In a broader comparative study, Eskandari et~al.~\cite{Eskandari2024} benchmarked eight deep learning architectures—including ResNet-50, DenseNet-121, ResNeXt-50, and Vision Transformers—on a large breast cancer patch dataset, finding that while Vision Transformers achieved the highest scores, ResNet-50 remained a strong and computationally efficient baseline.

Collectively, these studies highlight ResNet-50’s versatility and reliability as a backbone for histopathological image classification, making it an appropriate choice for the present work.

\subsection{Tiling}

Tiling, or subdividing microscopy images into smaller, uniformly sized patches, is a common strategy for increasing the number of training samples, improving focus on local features, and meeting memory constraints when working with high-resolution data. This approach has been applied across diverse microscopy modalities.
For example, Sharma and Yakimovich~\cite{Sharma2024} created a high-content microscopy dataset of sample preparation artifacts by extracting smaller image patches, facilitating supervised deep learning. In confocal microscopy, Vacca et~al.~\cite{Vacca2024} segmented high-throughput visual opsonophagocytosis (vOPA) assay images into multiple regions, implicitly enlarging the training set and improving classification performance. Ashesh et~al.~\cite{2022arXiv221112872A} introduced \textit{uSplit}, a patch-based framework designed for memory-efficient training, which partitions large microscopy images into context-preserving patches while minimizing tiling artifacts.

These strategies parallel the approach in this study, in which 1600$\times$1200 phase-contrast microscopy images were partitioned into non-overlapping grids. This increased the dataset size by a factor proportional to the grid resolution while preserving relevant cellular content for downstream morphology-based classification.

\section{Dataset}

The dataset we use is from~\cite{fletcher2025taxol}. It consists of grayscale ($1600 \times 1200$ pixels) phase-contrast microscopy images of C6 glioma cells, a rat-derived cell line commonly used as an \textit{in~vitro} model for glioblastoma. Cells were cultured for three days in dimethyl sulfoxide (DMSO) with either no drug (control) or paclitaxel (Taxol) at concentrations of 20, 40, or 100~nM, producing four experimental groups (109, 109, 110, and 110 images, respectively). Paclitaxel stabilizes microtubules, causing mitotic arrest and apoptotic cell death, leading to measurable morphological changes. 

\begin{figure}[htbp]
    \centering
    \begin{tabular}{cc}
        \includegraphics[width=0.38\columnwidth]{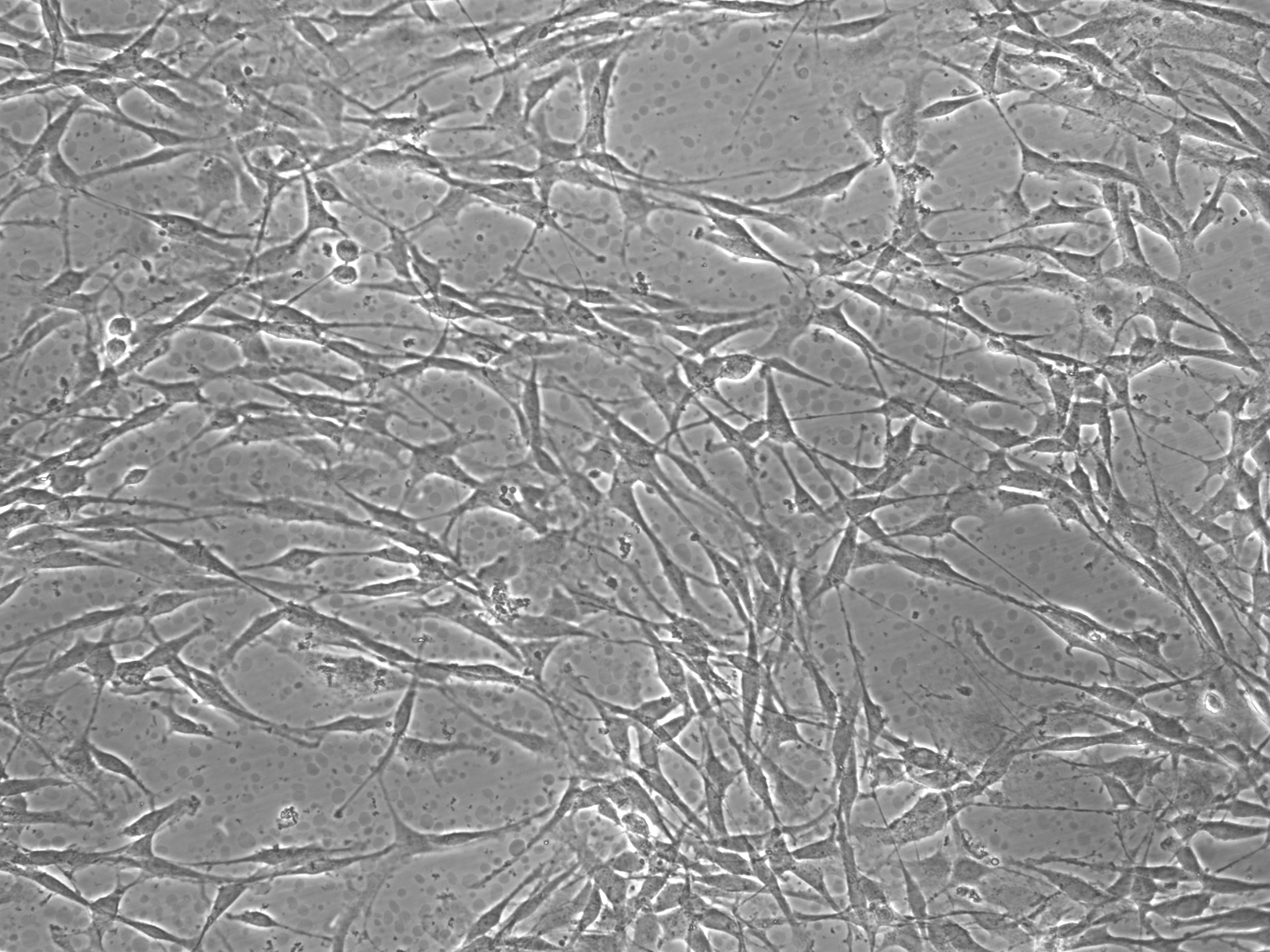} &
        \includegraphics[width=0.38\columnwidth]{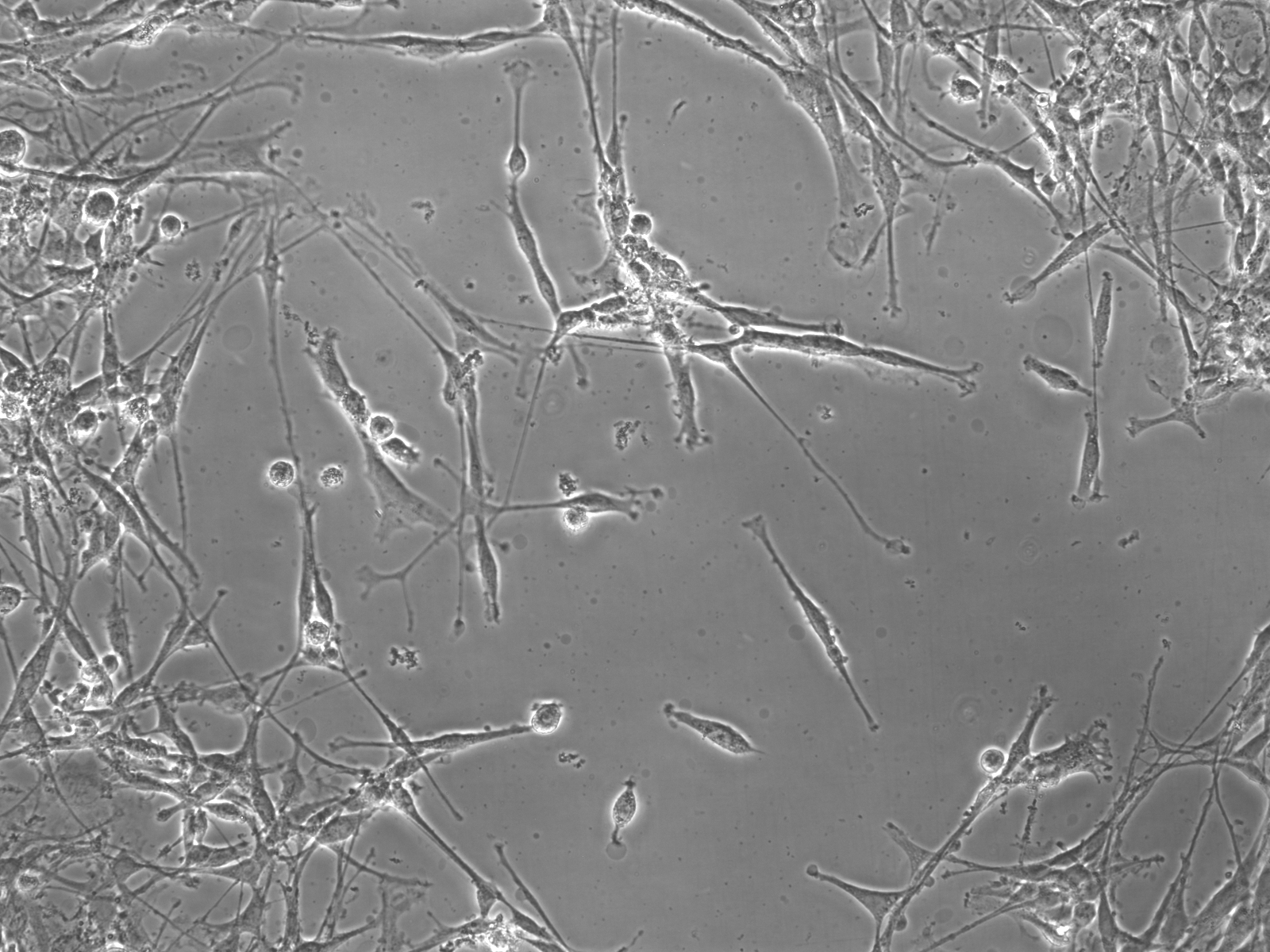} \\
        \small (a) Control & \small (b) 20~nM Taxol \\
        \includegraphics[width=0.38\columnwidth]{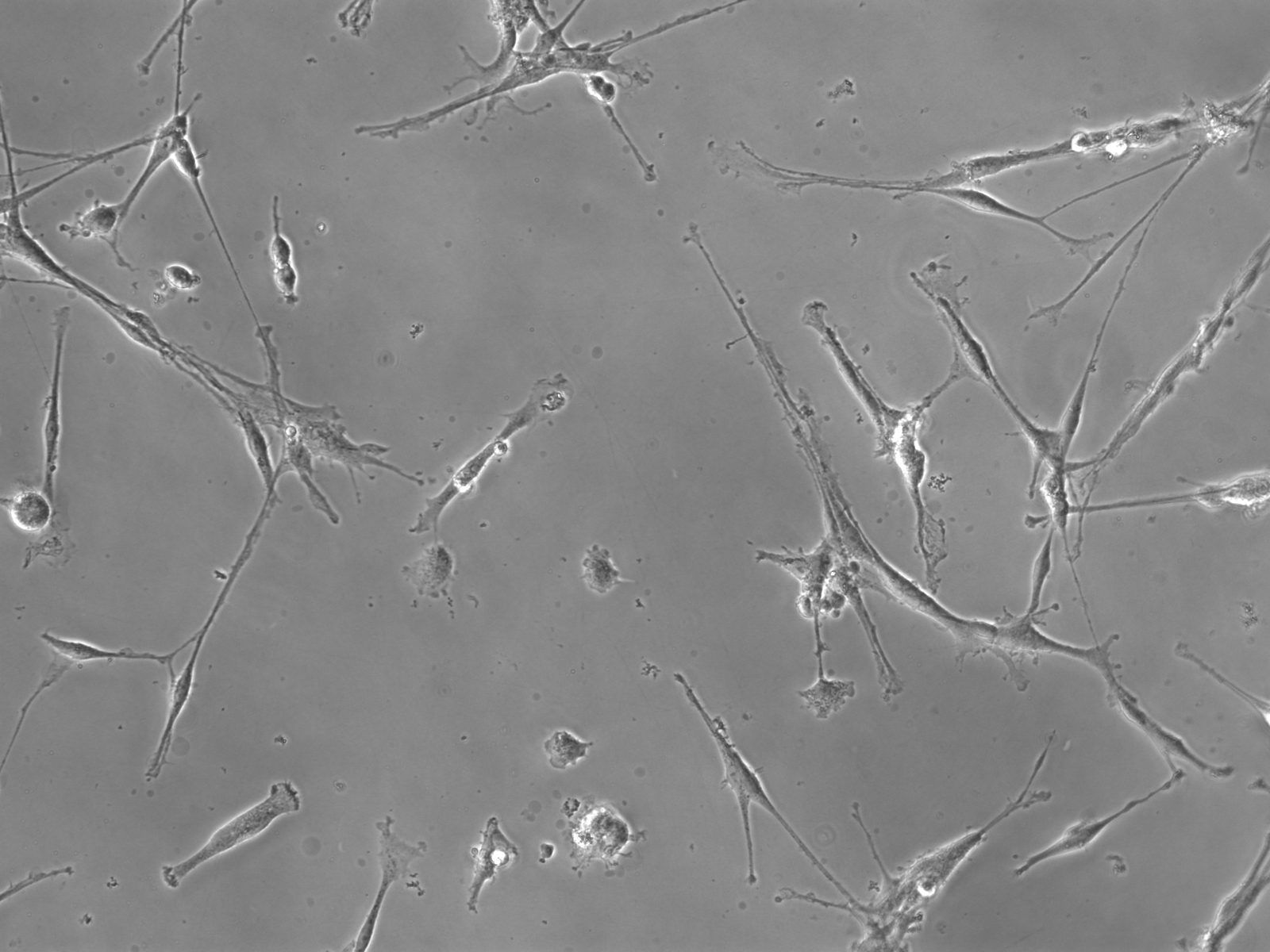} &
        \includegraphics[width=0.38\columnwidth]{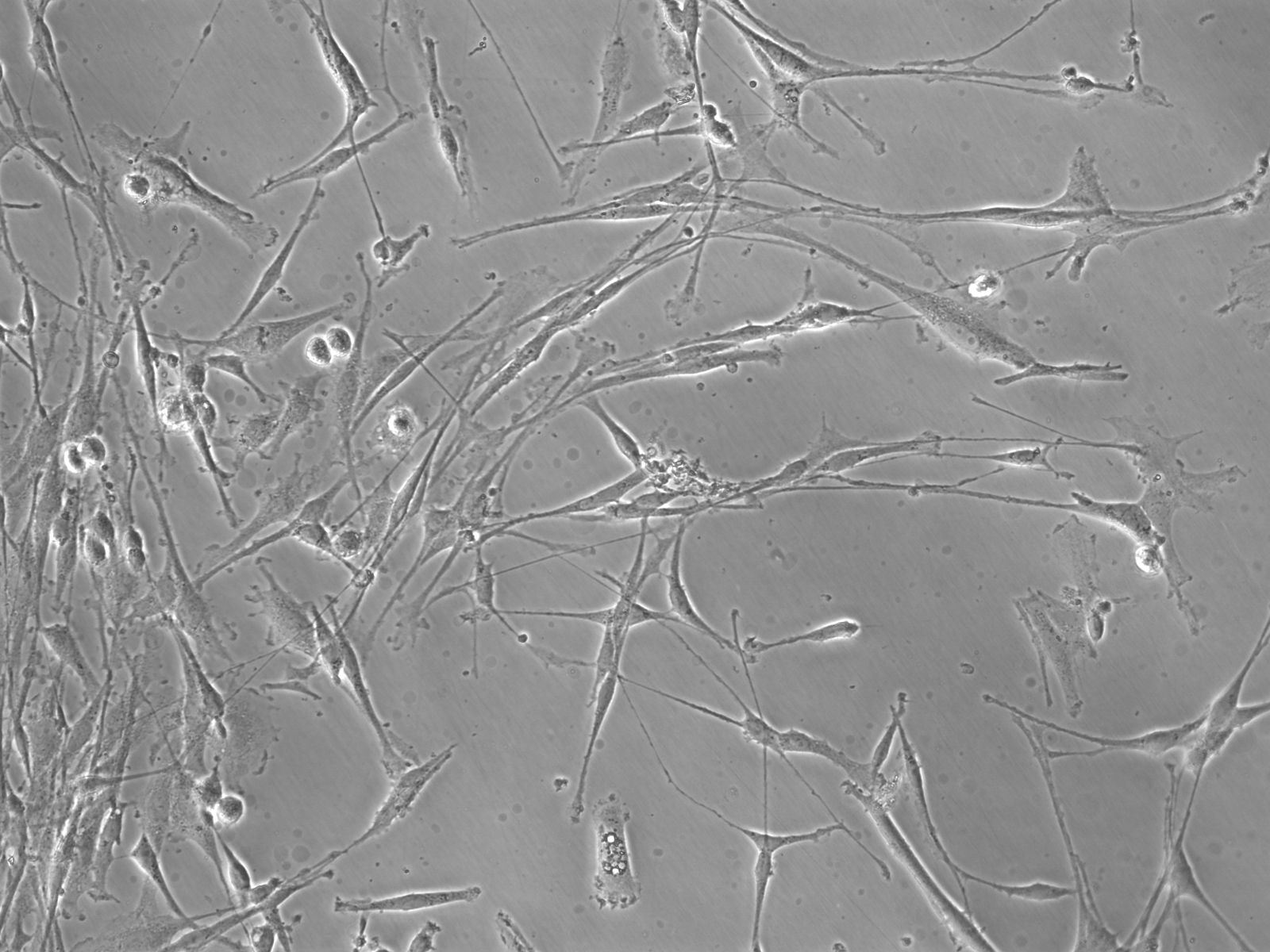} \\
        \small (c) 40~nM Taxol & \small (d) 100~nM Taxol \\
    \end{tabular}
    \caption{Representative phase-contrast microscopy images of C6 glioma cells from each experimental group.}
    \label{fig:dataset_examples}
\end{figure}

Fig.~\ref{fig:dataset_examples} shows one representative image from each group.
Table.~\ref{tab:dataset_splits} summarizes the class-balanced, stratified split: for each group, 16 images are used for validation and 16 for testing, with the remainder for training. Images were randomly shuffled before splitting and tracked by filename to prevent overlap across subsets, preserving class proportions and enabling fair comparison across methods.

\begin{table*}[t]
\caption{Image distribution across dataset splits by class (number of images).}
\label{tab:dataset_splits}
\centering
\renewcommand{\arraystretch}{1.2}
\setlength{\tabcolsep}{4pt}
\begin{tabularx}{0.8\textwidth}{|X|Y|Y|Y|}
\hline
\textbf{Class} & \textbf{Train} & \textbf{Val} & \textbf{Test} \\\hline
Control        & 77  & 16 & 16 \\ \hline
20~nM Taxol    & 77  & 16 & 16 \\ \hline
40~nM Taxol    & 78  & 16 & 16 \\ \hline
100~nM Taxol   & 78  & 16 & 16 \\ \hline
\textbf{Total} & 310 & 64 & 64 \\
\hline
\end{tabularx}
\end{table*}

\section{Methods}
In this section, we outline the pipeline for classifying Taxol exposure levels and provide the architectural and training details used in our experiments.

\begin{figure}[htbp]
    \centering
        \includegraphics[width=1.0\columnwidth]{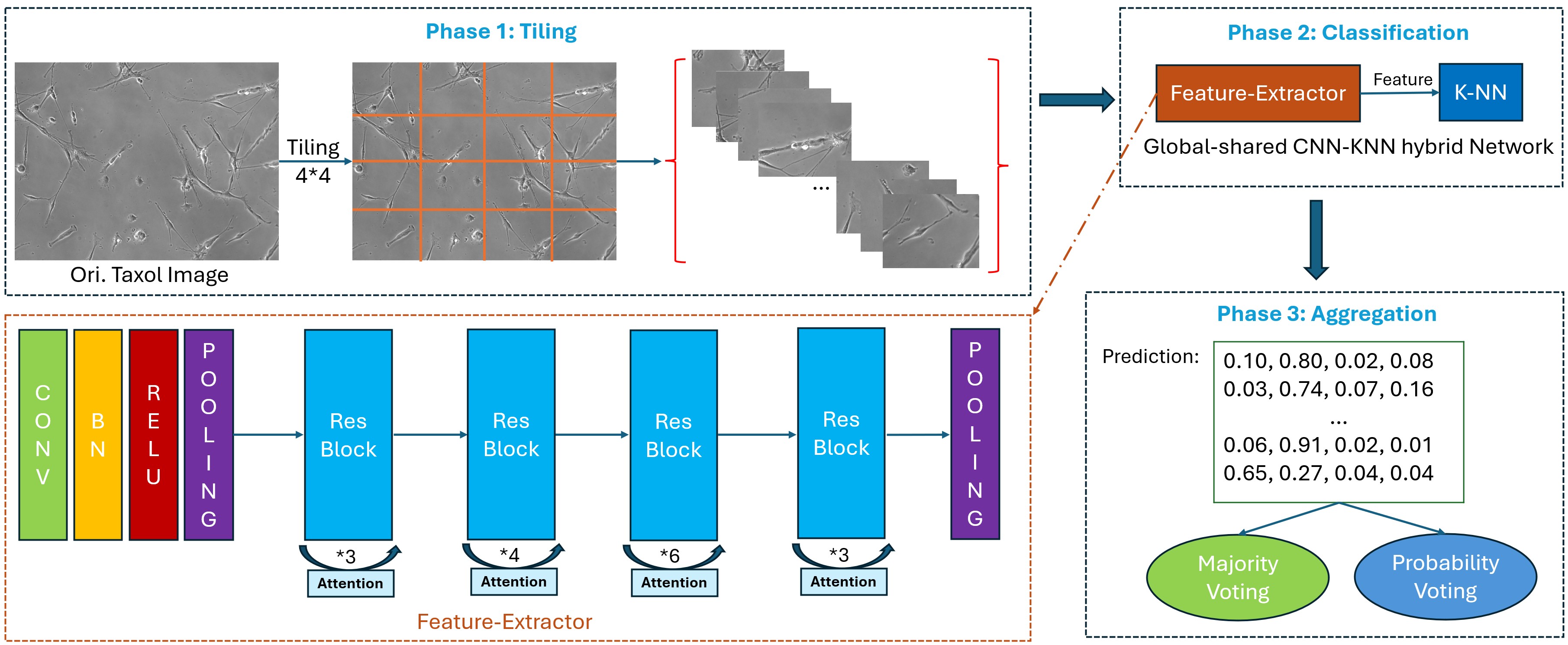} 
    \caption{Overall pipeline for paclitaxel exposure classification. Phase 1 tiles the image into an r-by-c grid. Phase 2 classifies each tile with a shared CNN and a K–NN head. Phase 3 aggregates tile outputs by majority or probability voting to obtain the image-level label.}
    \label{fig:pipe}
\end{figure}

Fig.~\ref{fig:pipe} shows a three-phase workflow. In \textbf{Phase 1}, each full-resolution image is partitioned into an  $r\times c$ grid of non-overlapping tiles. Tiles inherit the parent image label and are resized to the network input size. The grid size is treated as a hyperparameter. Then, all tiles are processed by a single, globally shared feature extractor. The resulting embeddings are classified with a k-nearest neighbors (K-NN) head, producing a per-tile four-class output in \textbf{Phase 2}. In \textbf{Phase 3}, tile-level predictions are aggregated into a single image-level label using either (i) \emph{majority voting} over tile class IDs or (ii) \emph{probability voting} that sums per-class scores across tiles,
$$
\hat{y} = \argmax_{c} \sum_{i=1}^{M} s_i[c].
$$
where $s_i[c]$ is the score (or probability) for class $c$ from tile $i$ and $M=r\cdot c$.

Details of our solutions are provided in the following subsections.

\subsection{Model}
The experiments employed ResNet-50, a convolutional neural network architecture introduced by He et~al.\ in 2015 \cite{he2016deep}. This network contains 50 layers organized into residual blocks, in which shortcut connections allow information to bypass certain layers. Such residual pathways reduce signal degradation during forward passes and improve gradient propagation in backpropagation, helping to overcome the vanishing gradient problem and enabling stable training of deeper models. Owing to these properties, ResNet-50 has become a standard backbone in a wide range of computer vision applications.
For this study, the architecture was adapted to accommodate grayscale microscopy images. The first convolutional layer was adjusted to accept single-channel input rather than the standard three-channel RGB format. The original 1000-class fully connected output layer was removed and replaced with a pair of dense layers: the first generated a 128-dimensional feature embedding, and the second produced the four-class output required for this classification task. The embedding layer served a dual purpose—supporting both end-to-end classification and downstream $k$-nearest neighbor ($k$-NN)~\cite{peterson2009k,kramer2013k} evaluation during validation and testing. All input images were resized to 224$\times$224 pixels to match the expected input dimensions of the ResNet-50 architecture.

% \subsection{Embedding}
% An embedding dimension of 128 was selected to balance feature compactness with sufficient representational capacity. Outputs from CNN backbones, such as the 2048-dimensional vectors produced by ResNet-50, often include redundant or noisy components and can encourage overfitting—especially in settings with limited training data. By constraining the embedding to 128 dimensions, the network is encouraged to retain only the most salient, task-relevant features, thereby improving generalization to unseen samples. Lower-dimensional, well-separated feature spaces are also advantageous for metric learning approaches, as distance-based classifiers such as $k$-nearest neighbors ($k$-NN) tend to operate more effectively under these conditions \cite{chen2020simple,khosla2020supervised}. Prior studies have reported that embeddings in the range of 128–256 dimensions strike an effective balance between discriminative capacity and overparameterization risk \cite{musgrave2020metric,schroff2015facenet}. The choice of 128 dimensions in this work reflects this evidence and supports both the classification and contrastive evaluation objectives of the study.

\subsection{Training and Evaluation}
All models were trained with multi-class cross-entropy using stochastic gradient descent (SGD) with momentum \(0.9\) and L2 weight decay \(5\times10^{-4}\). The backbone was ResNet-50 initialized with ImageNet-pretrained weights. Training ran for up to 200 epochs with early stopping when neither validation loss nor validation accuracy improved for 20 consecutive epochs. We saved a checkpoint whenever either metric improved, retaining the best-performing weights. Unless noted otherwise, we used a mini-batch size of 8 and a learning rate of 0.001, and we evaluated on the validation set at the end of each epoch.

To assess robustness beyond a single split, we performed five-fold cross-validation on the training set: data were partitioned into five equal folds; in each iteration, four folds were used for training and one fold served as a temporary validation set. The original validation split was not used during this process. Cross-validation was introduced after unexpectedly strong results on the fixed split and confirmed that performance was consistent across folds.

For testing, a unified pipeline supported both classification-head and embedding-based evaluation. For models with a softmax output, we forwarded test batches to obtain predicted labels (with logits and confidence scores) and reported accuracy as the proportion of correctly classified images. For embedding-based evaluation, we extracted 128-dimensional features from training and test images, fit a \texttt{KNeighborsClassifier} (Euclidean distance, \(k=5\)) on the training embeddings, and predicted test labels from the nearest neighbors. When evaluating tiled images, tile predictions were aggregated to the image level using either strict majority voting over tile class IDs or probability voting by averaging per-class softmax probabilities across tiles, which improved robustness and often increased full-image accuracy.

\subsection{Tiling}
\begin{figure}[htbp]
    \centering
    \includegraphics[width=0.5\textwidth]{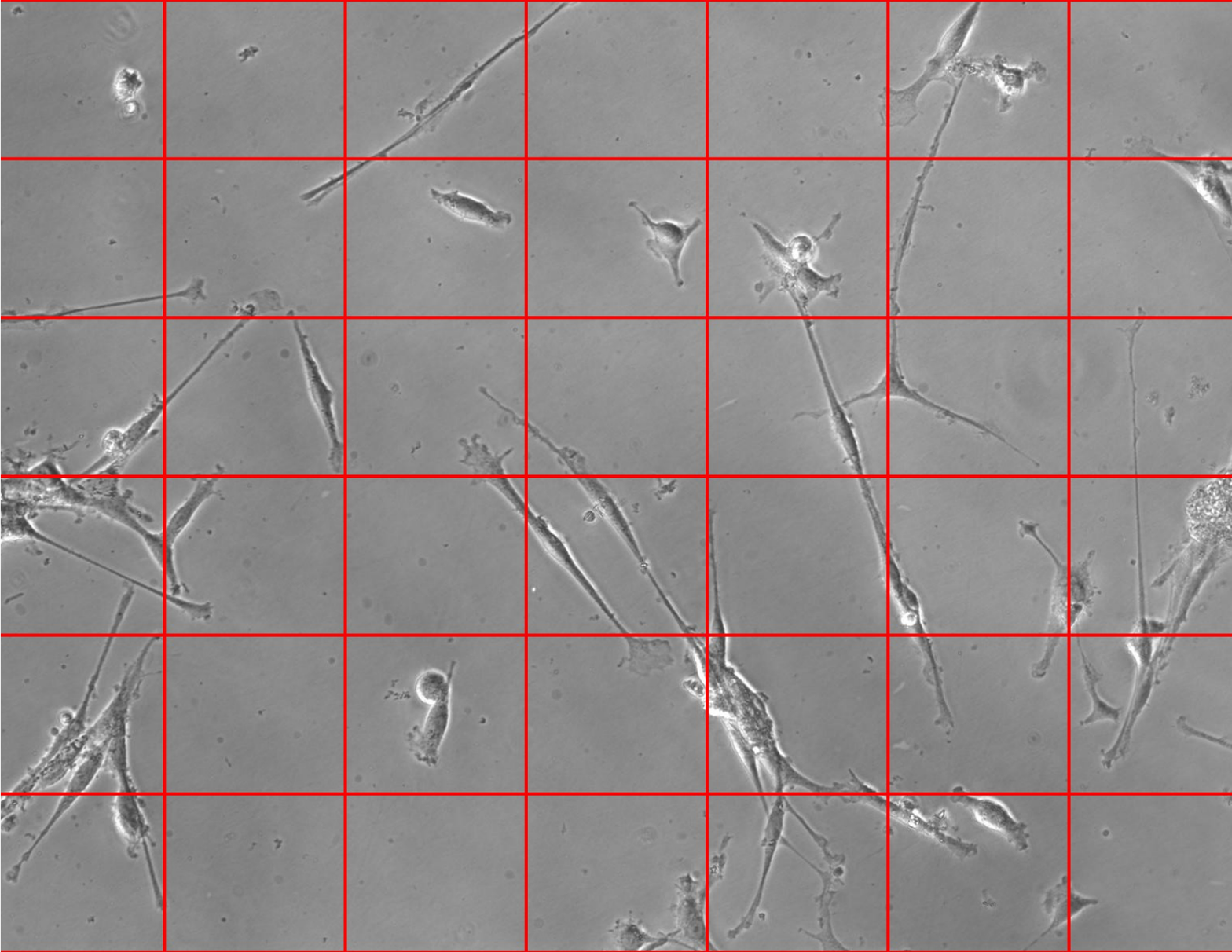}
    \caption{Example of tiling: 100\,nM class, image 068, divided into 6 rows by 7 columns.}
    \label{fig:100nm_068_6x7grid}
\end{figure}

The key preprocessing technique in this study was \textit{tiling}, the subdivision of microscopy images into smaller, uniformly sized sub-images. This approach has been widely applied in biomedical imaging to expand training datasets, reduce computational load, and enhance learning of fine-scale morphological features \cite{Sharma2024,Vacca2024,2022arXiv221112872A}. Tiling was implemented via a custom Python script that recursively traversed the dataset directory and split each image into a user-defined grid based on the specified number of rows and columns (Fig.~\ref{fig:100nm_068_6x7grid}). Pixels that did not fit evenly into the grid were discarded to ensure consistent tile dimensions. This step was important because all tiles were subsequently resized to 224$\times$224 pixels before being fed into the models, maintaining consistent scaling ratios across all inputs.

Tiling was performed \textit{after} the dataset had been split into training, validation, and test sets to prevent data leakage. Each tile was saved with a unique identifier encoding its spatial position within the original image, enabling traceability and supporting ensemble-style voting during evaluation. Tiles were then randomly shuffled and presented to the model as independent inputs, meaning the network learned from localized patches without access to the global spatial arrangement of the original image.

This approach increased the effective dataset size and was intended to help the models focus on localized morphological features of the cells—although this effect was not strongly supported by later results. Similar benefits have been reported in other microscopy domains, such as high-content HeLa cell imaging \cite{Sharma2024}, high-throughput phagocytosis assays \cite{Vacca2024}, and memory-efficient patch-based training in large-scale microscopy datasets \cite{2022arXiv221112872A}. Tiling was particularly suitable for the present dataset, in which cells were distributed randomly across the field of view, including corners, edges, and central regions, without a consistent spatial focus. However, this strategy may be less appropriate for datasets in which salient features are centrally located or tightly framed, as tiling could fragment important contextual information and impair classification performance. Systematic evaluation of different tiling configurations and their impact on classification accuracy became a major focus.

\section{Experiments}
In this section, we discuss the experimental setup and the resulting performance.

\subsection{Setup}
This study uses the benchmark dataset and evaluation protocol from \cite{fletcher2025taxol} and isolates the effect of tiling on classification performance. We evaluate a no-tiling baseline (full images) and a series of \(r\times c\) grids applied to the \(1600\times1200\) phase-contrast images:
\(1\times2\), \(2\times2\), \(2\times3\), \(3\times3\), \(3\times4\), \(4\times4\), \(4\times5\), \(5\times5\), \(5\times6\), \(6\times6\), \(6\times7\), \(7\times7\), \(7\times8\), \(10\times10\), and \(12\times12\).
This set spans coarse-to-fine granularity, enabling a controlled study of the trade-off between increased sample count (more tiles per image) and reduced global spatial context (smaller tiles). 
Table~\ref{tab:hyperparams} lists the hyperparameters used in all experiments. 

\begin{table}[htbp]
\centering
\caption{Summary of key hyperparameters used in model training.}
\label{tab:hyperparams}
\renewcommand{\arraystretch}{1.2}
\begin{tabular}{|l|l|}
\hline
\textbf{Hyperparameter} & \textbf{Value} \\\hline
Loss function         & Cross-entropy \\\hline
Optimizer             & SGD with momentum \\\hline
Momentum              & 0.9 \\\hline
Weight decay          & $5 \times 10^{-4}$ \\\hline
Learning rate         & 0.001 \\\hline
Batch size            & 8 \\\hline
Epochs (max)          & 200 \\\hline
Early stopping        & 20 epochs \\\hline
Embedding size        & 128 \\\hline
k in k-NN             & 5 \\\hline
Distance metric       & Euclidean \\
\hline
\end{tabular}
\end{table}

\subsection{Experimental Results}

\begin{table}[htbp]
\centering
\caption{Comparison with the baseline method on the Taxol dataset.}
\label{tab:bc}
\renewcommand{\arraystretch}{1.2}
\begin{tabularx}{0.8\linewidth}{|Y|Y|Y|Y|Y|}
\hline
\textbf{Model} & \textbf{Precision} & \textbf{Recall} & \textbf{F1} & \textbf{Acc.} \\
\hline
Ours & 0.97 & 0.97 & 0.97 & 0.97 \\
\hline
Baseline~\cite{fletcher2025taxol} & 0.76 & 0.75 & 0.75 & 0.75 \\
\hline
\end{tabularx}
\end{table}

\noindent\textbf{Compared with the baseline solution.}
The baseline for this Taxol dataset was proposed in \cite{fletcher2025taxol}. 
Table~\ref{tab:bc} summarizes the comparison between our approach and the baseline on the fixed test split. 
Our method improves all metrics—Precision, Recall, F1, and Accuracy—from $0.75{\sim}0.76$ to $0.97$. 
These gains are achieved by introducing tiling and aggregating tile predictions at the image level while keeping the backbone family and optimization setup consistent with the benchmark protocol. 
Overall, these results confirm that the proposed tiling-based pipeline is effective for Taxol classification on this benchmark.

\begin{figure}[htbp]
  \centering
  \includegraphics[width=\linewidth]{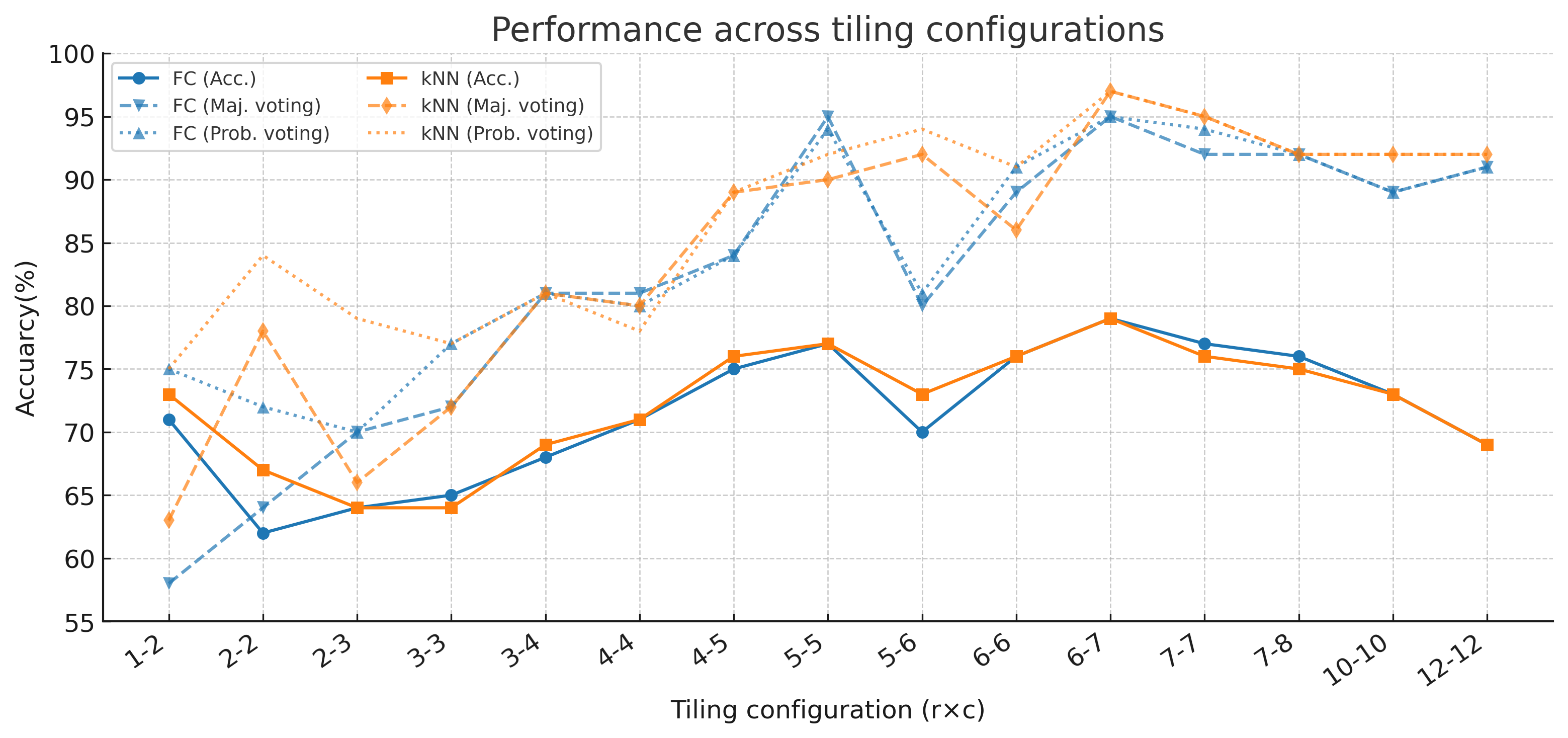}
  \caption{Accuracy across tiling configurations.}
  \label{fig:tile}
\end{figure}

\noindent
\textbf{Model performance across different tiling configurations.} Fig.~\ref{fig:tile} reports results across tiling grids. 
Here, \textit{FC} stands for the conventional CNN with a fully connected classification. 
\textit{kNN} stands for the hybrid in which the same CNN produces embeddings and labels are assigned by a kNN classifier in that embedding space. 
For the FC group, \textit{FC-Acc} is per-tile accuracy without aggregation, while \textit{FC-Maj} and \textit{FC-Prob} are image-level accuracies from majority voting and probability-weighted voting, respectively; \textit{kNN-Acc/Maj/Prob} are defined analogously for the kNN group.

Across tiling configurations, the CNN+KNN hybrid generally exceeds the traditional CNN with an FC head under both majority and probability voting.
In terms of tiling granularity, performance follows a “rise–peak–taper” trajectory: accuracy improves from coarse to mid-range grids, peaks at \(6{\times}7\) (FC \(95\%\), kNN \(97\%\)), and then declines for very fine splits. This pattern reflects a balance between benefits and costs: moderate tiling enlarges the effective sample size and isolates discriminative local structure, whereas excessive tiling erodes spatial context and increases the share of weakly informative tiles, reducing accuracy. We therefore adopt \(6{\times}7\) as the default split.

\section{Discussion}
To better understand how the model arrived at its predictions, we conducted a qualitative analysis of tile-level outputs from the 6×7 tiling configuration. Among all configurations tested, tiling consistently outperformed full-image training, confirming its effectiveness for this dataset. The improvement is likely due to tiling’s ability to increase the effective dataset size while forcing the model to focus on localized patterns rather than global image artifacts. The 6×7 configuration, paired with the model checkpoint corresponding to the highest validation accuracy, achieved the best performance on the test set and was therefore selected for in-depth exploration.
\begin{figure}[htbp]
  \centering
  \begin{subfigure}{0.26\linewidth}
    \centering
    \includegraphics[width=\linewidth]{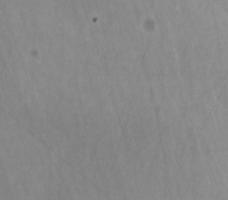}
    \caption{}
    \label{fig:tile_100nm_52_30_a}
  \end{subfigure}\hfill
  \begin{subfigure}{0.23\linewidth}
    \centering
    \includegraphics[width=\linewidth]{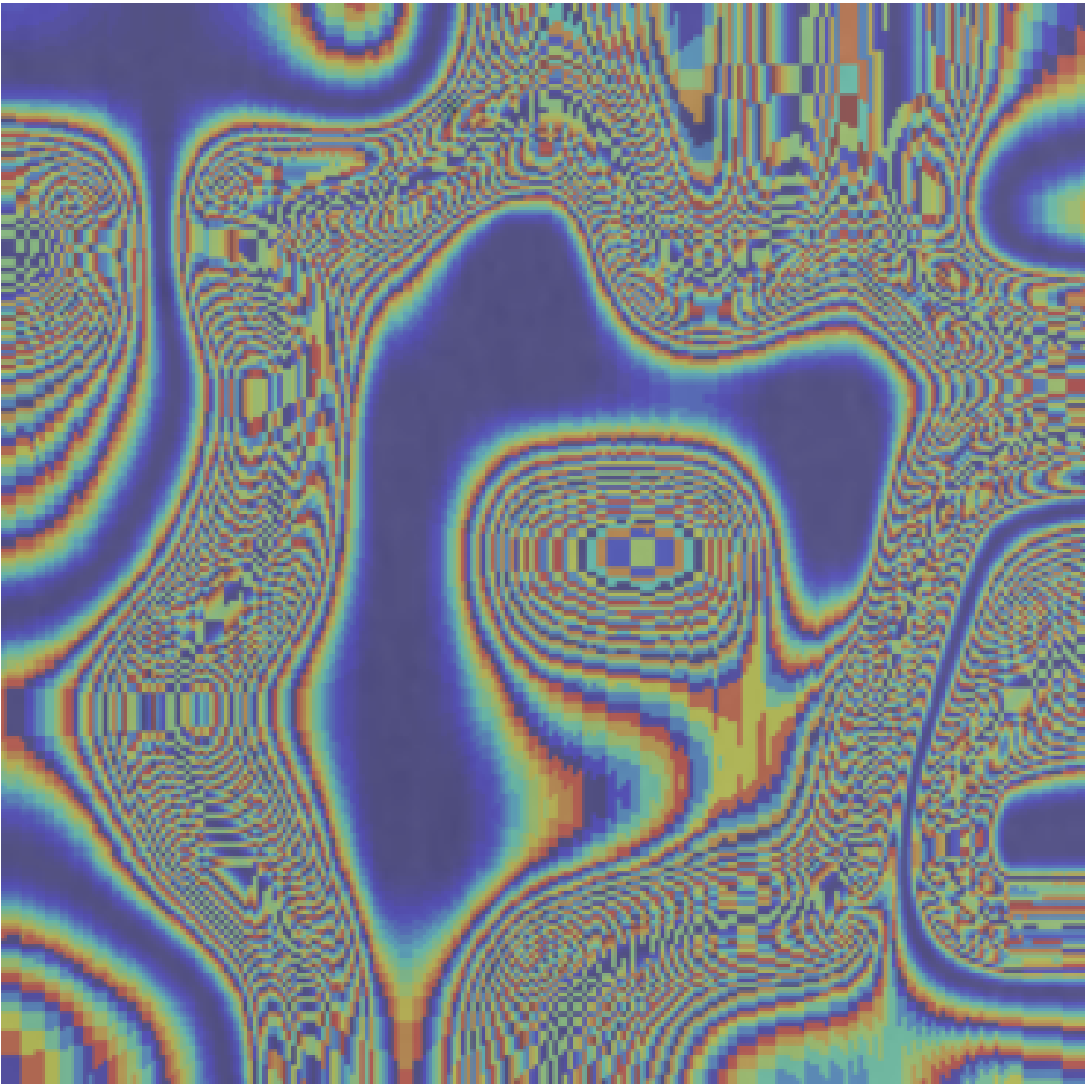}
    \caption{}
    \label{fig:heatmap_tile_100nm_52_30_b}
  \end{subfigure}\hfill
  \begin{subfigure}{0.26\linewidth}
    \centering
    \includegraphics[width=\linewidth]{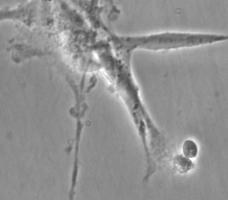}
    \caption{}
    \label{fig:tile_100nm_52_09_c}
  \end{subfigure}\hfill
  \begin{subfigure}{0.23\linewidth}
    \centering
    \includegraphics[width=\linewidth]{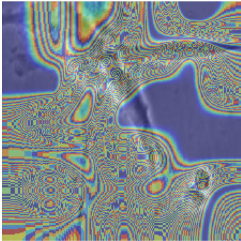}
    \caption{}
    \label{fig:heatmap_tile_100nm_52_09_d}
  \end{subfigure}
\caption{Example tiles and Grad-CAM maps from the 100\,nM class (Image 52). (a) Tile 30 (no visible cells). (b) Grad-CAM for (a). (c) Tile 09 (visible cells). (d) Grad-CAM for (c). In both cases, the most salient regions emphasize contextual textures rather than cellular morphology.}
  \label{fig:tiles_heatmaps_1x4}
\end{figure}
\subsection{Tile-Level Inference Beyond Cellular Morphology}
We observe that numerous tiles \emph{without visible cells} are nevertheless classified correctly and with high confidence, indicating reliance on cues beyond overt cellular morphology. To localize the evidence, we generated saliency maps using \textit{Grad-CAM}~\cite{selvaraju2017grad} and \textit{Score-CAM}~\cite{wang2020score} for both cell-absent and cell-present tiles. In both cases, the most activated regions rarely overlap with cells, instead emphasizing fine-scale, field-wide textures and contextual patterns. The agreement between these two methods supports the conclusion that predictions are driven primarily by non-cellular cues. For illustration, Fig.~\ref{fig:tiles_heatmaps_1x4} exhibites saliency concentrated on contextual textures rather than cellular morphology. Here we only display Grad-CAM for consistency, and Score-CAM produces qualitatively similar results, so for limited space, we donot show Score-CAM result here. The heat maps were generated from the resized images used as model inputs and are therefore square, whereas the original tiles retain the rectangular aspect ratio of the source images.

\begin{figure*}[t]
  \centering
  \begin{subfigure}{0.48\textwidth}
  \centering
    \includegraphics[width=0.8\linewidth]{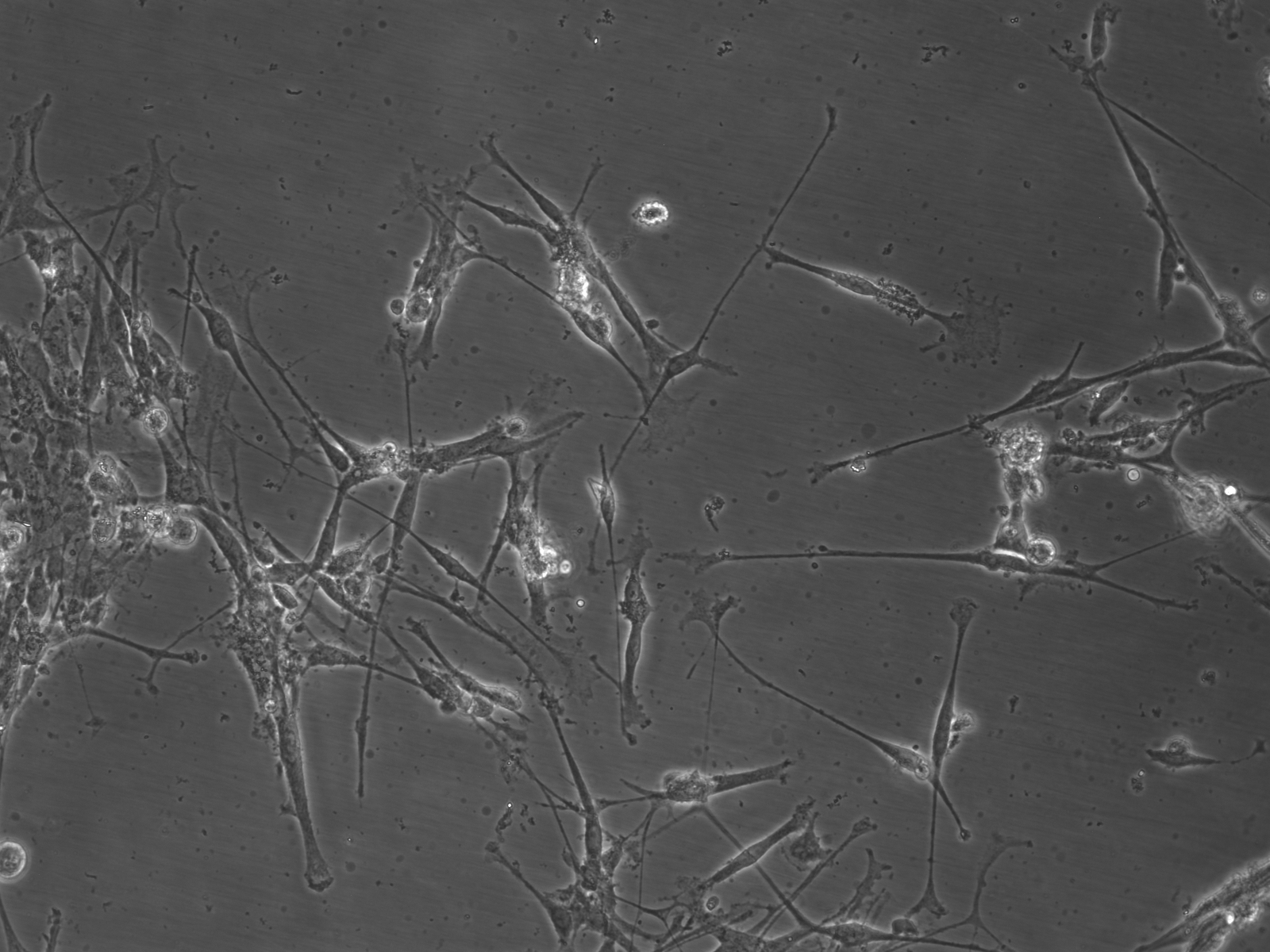}
    \caption{Full image (100\,nM, Image 006)}
  \end{subfigure}\hfill
  \begin{subfigure}{0.48\textwidth}
  \centering
    \includegraphics[width=0.6\linewidth]{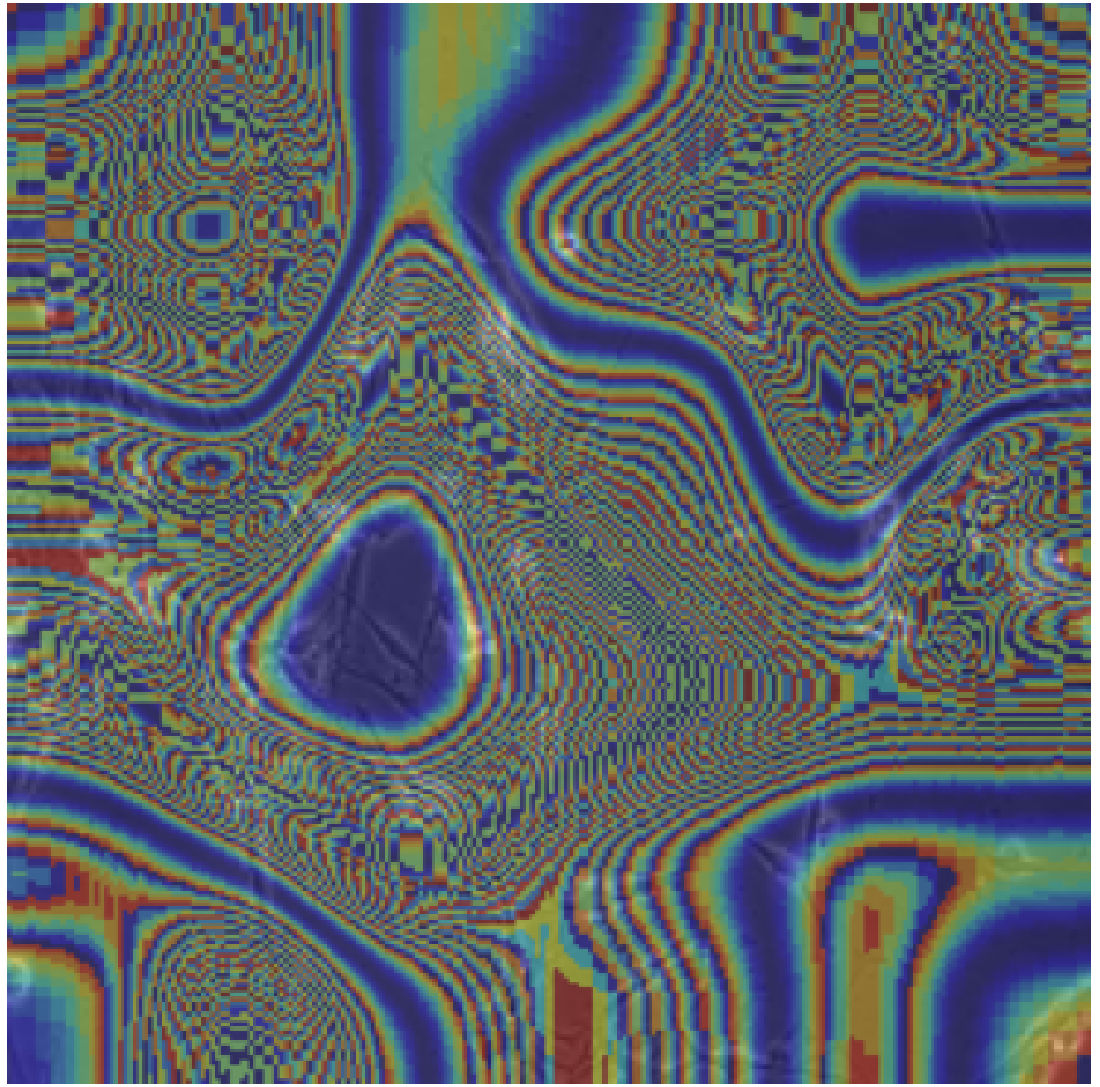}
    \caption{Grad-CAM for Image 006}
  \end{subfigure}
  \caption{Full 100\,nM Image 006 and its Grad-CAM heatmap; activation does not correspond to cells.}
  \label{fig:full_image_and_heatmap_100nm_006}
\end{figure*}

We use the same method for the full iamge analysis, which corroborates the same phenomenon. When the model is applied to full, untiled images, high-confidence predictions persist while saliency concentrates on contextual microtextures rather than on cellular contours (see full-image example and heat map in Fig.~\ref{fig:full_image_and_heatmap_100nm_006}). The consistency across tile-level and full-image settings—and across both FC and kNN evaluators—indicates that these contextual cues are stable signals in this dataset rather than artifacts of tiling or classifier choice.

Taken together, these findings highlight that tiling improves performance markedly on this benchmark (peaking at \(6{\times}7\)), yet the attention maps rarely align with cellular morphology. A plausible explanation is that the classifier exploits peri-cellular/contextual microtextures that co-vary with exposure and are stable at the tile scale. While this yields strong in-sample accuracy, it raises a central concern: \textit{robustness} under distribution shift (e.g., different plates, optics, illumination, or laboratories). Accordingly, evaluation should pair quantitative metrics with qualitative interpretability to assess transparency and biological relevance. This concern also points to concrete directions for computer-aided cell-structure classification: consistent with human intuition that morphology should drive decisions, systematically evaluate—and, where feasible, enforce—overlap between model attention and cellular structures; mitigate the influence of non-cellular cues through acquisition/preprocessing controls; and align saliency with domain-expert priorities. These steps aim to improve robustness while preserving the efficiency gains delivered by tiling.

\section{Conclusion}
This study demonstrates both the potential and the pitfalls of applying deep learning to phase-contrast microscopy. A simple tiling-and-aggregation pipeline solves the Taxol classification problem on this benchmark, reaching 95–97\% test accuracy and surpassing the prior baseline by ~20 percentage points. At the same time, interpretability analyses show that predictions often rely on contextual, non-cellular cues rather than overt cellular morphology. Thus, while tiling is highly effective for this dataset, the learned cues raise concerns about robustness under distribution shift. Reporting both the quantitative gains and the saliency evidence makes clear what the model is using—and where caution is warranted.


\bibliographystyle{splncs04}
% \bibliography{ref}

\end{document}